\documentclass{article}
\usepackage{iclr2027_conference,times}

\usepackage{geometry}
\usepackage{amsmath,amsfonts,bm}

\def\eqref#1{equation~\ref{#1}}

\def\1{\bm{1}}

\def\rvepsilon{{\mathbf{\epsilon}}}

\def\va{{\bm{a}}}

\def\mI{{\bm{I}}}

\DeclareMathAlphabet{\mathsfit}{\encodingdefault}{\sfdefault}{m}{sl}
\SetMathAlphabet{\mathsfit}{bold}{\encodingdefault}{\sfdefault}{bx}{n}

\usepackage{amsmath,amssymb}
\usepackage{booktabs}
\usepackage{graphicx}
\usepackage{multirow}
\usepackage{subcaption}
\usepackage{tabularx}
\usepackage{xcolor}
\usepackage{colortbl}
\usepackage{siunitx}
\usepackage{marvosym}
\usepackage{tikz}
\usetikzlibrary{arrows.meta,calc,positioning,shapes.geometric,fit,backgrounds,decorations.pathreplacing,patterns}
\usepackage{microtype}
\usepackage{wrapfig}
\usepackage{needspace}
\usepackage[most]{tcolorbox}
\usepackage[scaled=0.98]{helvet}
\usepackage{hyperref}
\usepackage{url}

\definecolor{grouplab}{HTML}{6E6E73}
\definecolor{oursrow}{HTML}{F2EFF7}
\definecolor{muted}{HTML}{666666}
\newcommand{\ours}{\rowcolor{oursrow}}

\newcommand{\cmark}{\ensuremath{\checkmark}}
\newcommand{\pmark}{\ensuremath{\times}}

\definecolor{undcolor}{HTML}{78669A}
\definecolor{gencolor}{HTML}{9C7BB5}
\definecolor{jointcolor}{HTML}{D95F59}

\definecolor{prelimcol}{HTML}{B8541A}

\newlength{\sbarw}
\newcommand{\sharebar}[2]{%
  \begin{tikzpicture}[baseline=-0.2ex,x=\sbarw,y=1ex]
    \fill[muted!10] (0,0) rectangle (1,1.15);
    \fill[#2] (0,0) rectangle ({#1},1.15);
    \draw[muted!40,line width=0.2pt] (0,0) rectangle (1,1.15);
  \end{tikzpicture}}

\definecolor{findingblue}{RGB}{45,82,135}
\definecolor{findingbg}{RGB}{240,246,255}
\newtcolorbox{findingbox}{
    colback=findingbg,
    colframe=findingblue,
    boxrule=0.8pt,
    arc=3mm,
    left=7pt,
    right=7pt,
    top=5pt,
    bottom=5pt,
    boxsep=0pt,
    before skip=6pt,
    after skip=6pt
}
\newcommand{\shows}[1]{%
  \begin{findingbox}
    \textcolor{findingblue}{\textbf{Finding.}} #1
  \end{findingbox}
}

\newcommand{\sysname}{NEO-VLA}

\newcommand{\papersecref}[1]{Section~\ref{#1}}
\newcommand{\paperfigref}[1]{Figure~\ref{#1}}
\newcommand{\papertabref}[1]{Table~\ref{#1}}

\definecolor{titleboxbg}{HTML}{F3F0FA}
\definecolor{titleboxborder}{HTML}{CFC4E8}
\definecolor{titleink}{HTML}{2E2A36}
\definecolor{linkviolet}{HTML}{6E59B3}
\definecolor{urlmagenta}{HTML}{C93D8E}
\hypersetup{
    colorlinks=true,
    linkcolor=linkviolet,
    citecolor=linkviolet,
    urlcolor=urlmagenta,
    pdftitle={From Pixels to Actions -- Towards Native Vision-Language-Action Policies},
    pdfauthor={Weichen Fan and coauthors}
}
\iclrfinalcopy

\begin{document}
\fancyhead{}
\renewcommand{\headrulewidth}{0pt}

\begin{tcolorbox}[
    colback=titleboxbg,
    colframe=titleboxborder,
    boxrule=0.55pt,
    arc=8mm,
    width=\textwidth,
    left=20pt,
    right=20pt,
    top=18pt,
    bottom=16pt,
    before skip=0pt,
    after skip=14pt
]
\color{titleink}

{\color{black}\sffamily\bfseries\fontsize{19}{23}\selectfont
From Pixels to Actions -- Towards\\
Native Vision-Language-Action Policies\par}
\vspace{0.80em}

{\color{black}\sffamily\bfseries\fontsize{9}{11.3}\selectfont\raggedright
Weichen Fan\textsuperscript{1,*},\quad
Haiwen Diao\textsuperscript{1,*,\ensuremath{\sharp}},\quad
Penghao Wu\textsuperscript{1},\quad
Siyuan Hu\textsuperscript{1},\quad
Ziqi Huang\textsuperscript{1},\quad
Tianqi Liu\textsuperscript{1,3},\quad
Mingze Sun\textsuperscript{1},\quad
Ziyi Wang\textsuperscript{1},\quad
Zhongang Cai\textsuperscript{2},\quad
Hanming Deng\textsuperscript{2},\quad
Quan Wang\textsuperscript{2},\quad
Lei Yang\textsuperscript{2},\quad
Lewei Lu\textsuperscript{2},\quad
Fangzhou Hong\textsuperscript{3},\quad
Zhaoxi Chen\textsuperscript{3},\quad
Dahua Lin\textsuperscript{2},\quad
Ziwei Liu\textsuperscript{1,\ensuremath{\dagger}}\par}
\vspace{0.45em}

{\fontsize{10.2}{12.4}\selectfont\raggedright
\textsuperscript{1}S-Lab, Nanyang Technological University,\quad
\textsuperscript{2}SenseTime Research,\quad
\textsuperscript{3}Ropedia\par}
\vspace{0.2em}

{\fontsize{8.4}{10.2}\selectfont
\textsuperscript{*}Equal contribution.\quad
\textsuperscript{\ensuremath{\sharp}}Project Lead.\quad
\textsuperscript{\ensuremath{\dagger}}Corresponding author.\par}
\vspace{0.70em}

\begingroup
\fontsize{10.5}{13.2}\selectfont
\setlength{\parindent}{0pt}
\setlength{\parskip}{4pt plus 1pt minus 1pt}
\renewenvironment{abstract}{\par\noindent\ignorespaces}{\par}
\begin{abstract}

Vision-language-action models are typically built on a simple premise: perception is first compressed by visual encoders, and action reasoning begins only afterwards. Yet manipulation often hinges on precisely the visual signals most vulnerable to this abstraction, e.g., local geometry, boundaries, depth, and subtle state changes. 
This raises a fundamental question: \emph{should action policies inherit a fixed visual endpoint, or directly access evolving visual computation?}
We introduce \textbf{\sysname{}}, a structurally encoder-free vision-language-action framework. Images are tokenized by lightweight patch convolutional layers and evolve jointly with language throughout the backbone, allowing a flow-matching action expert to draw from instruction-conditioned visual representations at different depths.
This simple change reveals a consistent pattern. Native architectures outperform encoder-based counterparts under controlled comparisons; full fine-tuning reduces but does not erase the gap, and additional scale alone cannot recover it. More strikingly, layer-wise probing shows that the action-critical signals often peak well before the final visual representation. Across diverse manipulation settings and distribution shifts, \sysname{} consequently achieves stronger and more robust control. Together, these results suggest that embodied control benefits from access to visual representations across the computational hierarchy, rather than a fixed encoder output.

\end{abstract}

\endgroup

\vspace{0.5em}
{\sffamily\bfseries\fontsize{8}{11}\selectfont
Correspondence:\quad
\url{ziwei.liu@ntu.edu.sg}\par}

\vspace{0.3em}
{\sffamily\bfseries\fontsize{8}{11}\selectfont
Project Page:\quad\url{https://www.lmms-lab.com/posts/neo_vla/}\par}
\vspace{0.2em}
{\sffamily\bfseries\fontsize{8}{11}\selectfont
GitHub Repo:\quad\url{https://github.com/EvolvingLMMs-Lab/NEO}\par}
\vspace{0.2em}
{\sffamily\bfseries\fontsize{8}{11}\selectfont
Model Weight:\quad\url{https://huggingface.co/weepiess2383/NEO-VLA}\par}
\vspace{0.2em}
{\sffamily\bfseries\fontsize{8}{11}\selectfont
Dataset Link:\quad\url{https://huggingface.co/datasets/weepiess2383/NEO-VLA-Action-Pretrain}\par}

\end{tcolorbox}

\section{Introduction}
\label{sec:intro}

Robotic manipulation demands both semantic understanding and fine-grained spatial reasoning. A policy must identify \emph{what} an instruction refers to, infer \emph{where} the target lies, and determine \emph{how} the gripper should move relative to it. This raises a fundamental question for vision-language-action (VLA) models: should action prediction rely only on the final visual representation, or also on the intermediate visual computation?

Most existing VLAs make this choice implicitly. Images are first compressed by pretrained vision encoders, such as CLIP, SigLIP, or DINOv2~\citep{clip2021,siglip2023,dinov2_2023}, before the resulting features are passed to the language backbone and action model~\citep{rt2_2023,openvla2024,pi0_2024,grootn1_2025}. 
This interface exposes the encoder output while hiding the visual hierarchy that produced it.
Growing evidence suggests that this separation may constrain embodied control: visual processing is critical to VLA performance~\citep{vlm4vla2026}, visual components are disproportionately sensitive to pruning~\citep{zhang2026redundancy}, and action fine-tuning can degrade pretrained visual representations~\citep{kachaev2025blind}. 
We instead reconsider the interface itself, enabling action prediction to directly exploit the evolving visual representations within the backbone.

\begin{figure}[t]
\centering
\includegraphics[width=\textwidth]{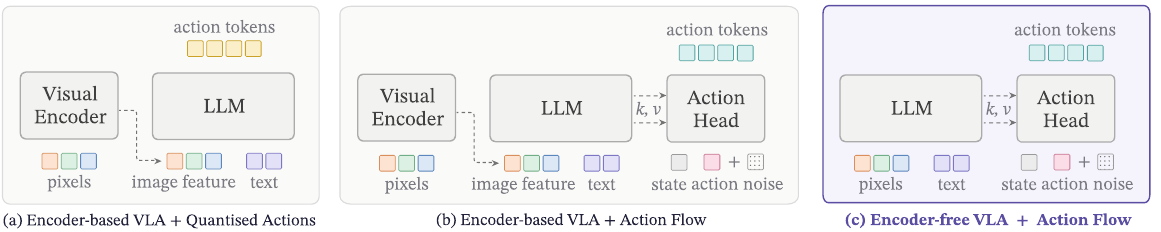}
\caption{%
\textbf{Three VLA designs}: encoder-based vs. encoder-free backbones, and discrete vs. continuous actions.
\textbf{(a)} A separate visual encoder converts pixels into image features, which are consumed by the language backbone together with text tokens; actions are generated as discrete tokens.
\textbf{(b)} The same encoder-based visual interface is retained, while a separate action head predicts continuous actions conditioned on the backbone.
\textbf{(c)} \sysname{} removes the separately pretrained visual encoder: a lightweight patch embedder maps pixels directly into the backbone token space, while text tokens enter the backbone jointly. A narrow action expert then reads cached keys and values from selected backbone layers and predicts a flow-matching velocity field.}
\label{fig:designspace}
\end{figure}

We introduce \sysname{}, a structurally native VLA policy built directly on pixels and words, requiring no separately pretrained vision encoder, image codebook, or autoencoder.
Following encoder-free models~\citep{fuyu-8b,eve2024,diao2025EVEv2,neo,neoov2026}, a lightweight convolutional patch embedder maps raw pixels into the backbone token space, where image patches and instruction words are processed jointly within a single holistic architecture in \paperfigref{fig:designspace}. A flow-matching action expert then reads cached keys and values from diverse backbone layers with different representation hierarchies to predict continuous action chunks. 

This design makes visual readout depth an explicit policy choice: the action expert can draw from early, late, or multiple stages of visual computation without introducing a separate visual pathway. In conventional encoder-based VLAs, by contrast, accessing different language-backbone layers still places the policy downstream of a fixed visual endpoint. \sysname{} thus enables a direct investigation of where action-relevant visual information emerges and which stages of visual computation are most valuable for control.

Our analysis reveals a clear depth-dependent pattern. With the backbone frozen, readouts that include early layers consistently outperform those restricted to later representations; after full fine-tuning, combining early and late features performs best. Targeted interventions lead to the same conclusion: swapping early-layer activations across observations induces substantially larger prediction errors than perturbing later layers. Together, these results indicate that action-critical information emerges well before the final representations, when fine-grained spatial and geometric cues remain more directly accessible. Preserving access to these intermediate computations therefore provides a strong basis for precise and generalizable control.

We further compare \sysname{} with Qwen3-VL-2B and Gemma-3-4B, which offer comparable understanding capabilities, using the same action expert and robot-training corpus. \sysname{} consistently achieves higher LIBERO success rates under both frozen-backbone and full-finetuning settings, with the advantage particularly pronounced when the backbone is frozen and policy learning must rely directly on pretrained visual representations. Although differences in pretraining and model capacity preclude attributing the gains solely to encoder removal, these results establish the native backbone as a strong and transferable foundation for action learning.

We then evaluate \sysname{} on LIBERO, LIBERO-plus, and RoboTwin 2.0, covering both standard manipulation and challenging distribution shifts in visual appearance, language, and environment. \sysname{} achieves strong in-distribution performance while maintaining robust control under these shifts, with particularly pronounced gains under camera perturbations and image noise. These results complement the layer-wise analysis by showing that preserving access to intermediate visual computation is not only useful in controlled ablations, but also translates into practical robustness across diverse manipulation settings. Taken together, our findings point to a broader principle for embodied intelligence: effective control depends not only on what the backbone represents, but also on which stages of its visual computation remain accessible to action.

Our main contributions are summarised as follows.
\begin{itemize}
  \item \textbf{A structurally native VLA policy.}
  We introduce \sysname{}, which combines a lightweight pixel-to-token embedder,
  a shared vision-language backbone, and a flow-matching action expert with
  configurable access to backbone layers with diverse visual hierarchies.

  \item \textbf{A systematic empirical study of action readout depth.}
  Layer-map comparisons with matched expert capacity, together with
  attachment-point interventions, show that early visual representations
  are pretty effective for control, particularly when the backbone is frozen.

  \item \textbf{Evaluation across backbone families and manipulation benchmarks.}
  We compare native and encoder-based backbones using the same action-expert
  design and robot-training corpus, and show strong performance on
  LIBERO, LIBERO-plus, and RoboTwin 2.0.
\end{itemize}
\section{Related Work}
\label{sec:related}

\paragraph{Vision-Language-Action Models.}
Existing VLA policies typically attach an action decoder to a pretrained vision-language
model. Early systems emit actions as discrete text tokens
\citep{rt2_2023,openvla2024}, while later work uses continuous diffusion or
flow-matching decoders over backbone features
\citep{pi0_2024,pi05_2025,grootn1_2025}; \citet{octo2024} instead trains a
transformer policy from scratch with a diffusion head.
Our design differs along two axes:
whether the visual pathway introduces a separately pretrained vision encoder, and whether
actions are represented discretely or continuously.
Most prior work has focused on the action side while leaving the visual pathway
largely unchanged.
Images are commonly processed by pretrained encoders such as CLIP, SigLIP, or
DINOv2 \citep{clip2021,siglip2023,dinov2_2023}, often at a fixed resolution and
sometimes with the encoder frozen during policy training.
Recent studies suggest that this front end can limit control:
\citet{vlm4vla2026} identify the visual module as a major bottleneck,
\citet{zhang2026redundancy} find the vision encoder, language backbone, and
projector among the components most sensitive to pruning, and
\citet{kachaev2025blind} show that action finetuning can degrade visual
representations.
Recent unified systems also tend to preserve the visual front end, adding
modality-specific experts for generation, foresight, or world modelling above a
shared attention stream
\citep{mot2025,internvla_a1_2026,internvla_a15_2026,f1_2025,eo1_2025,
bagelvla2026,rynnvla002_2025,mmact2025,mmadavla2026}.
\citet{uam2026} instead adds a second visual pathway, arguing that one encoder
may not serve both language grounding and control well.
Our focus is different: we vary the visual pathway itself while keeping the action expert, the layers it reads, the training data.

\paragraph{Native Multimodal Models.}
A separate line of work removes the vision encoder itself, mapping pixels through a lightweight learned patch embedder directly into the language backbone
\citep{fuyu-8b,eve2024,diao2025EVEv2,monointernvl2025,neo,neoov2026,
sensenova2026sensenovau1,diao2026sensenovau2,thinkingmachines2026interactionmodels}.
In this literature, \emph{native} can refer to different ideas: removing a separate vision module, training multimodally from the start
\citep{tong2026beyond,han2026towards}, or preserving the backbone's original
format and objectives during policy training \citep{internvla_a15_2026}.
We use the term only in the structural sense: no separately trained vision
module, image codebook, or autoencoder.
Control models have evolved along related lines. \citet{univla2025} represent vision, language, and action as unified discrete token sequences, and \citet{dynamicvla2026} replace the transformer-based vision encoder with a convolutional alternative.
Concurrent work studies understanding and generation without a separately
trained vision module \citep{wu2026synergy}.
Our focus is narrower: we compare visual pathways under a matched action expert
and study which backbone depths the expert should read.

\section{Model}
\label{sec:model}

\begin{figure}[t]
  \centering
  \includegraphics[width=\textwidth]{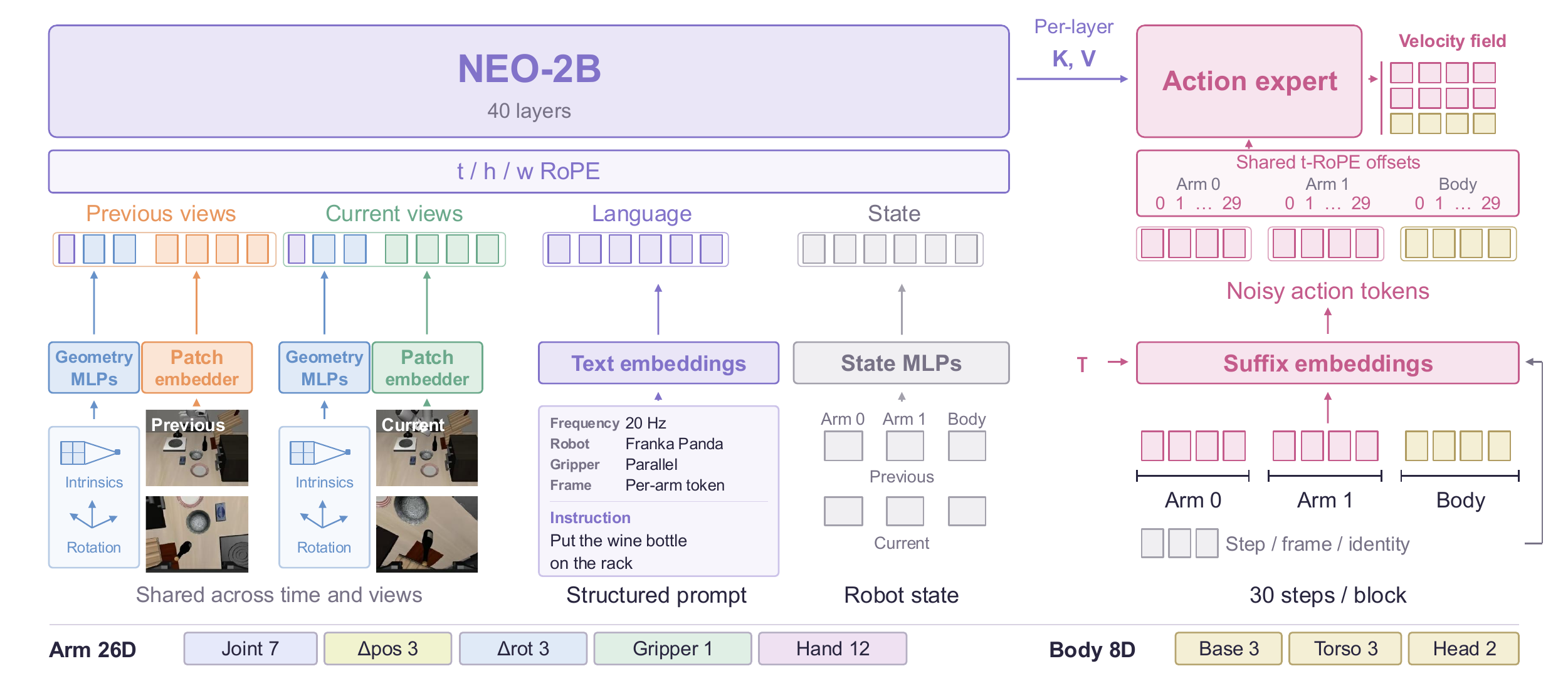}
\caption{\textbf{\sysname{}: pixels in, continuous actions out.} Previous and current multi-view observations preserve their aspect ratios within a fixed token budget and enter patch embedder shared across time and views. The visual path uses no vision transformer, autoencoder, or vector-quantised codebook. Shared MLPs embed camera intrinsics and rotations, while state MLPs embed previous and current robot states. These tokens join a
structured prompt describing control frequency, embodiment, frame reference, and task instruction.
A separate, narrow, randomly initialised action expert jointly attends to cached prefix keys and values from selected backbone layers and its own action-token keys and values. Its suffix embeddings combine projected noisy actions and flow time $\tau$ with additive step, frame, and identity embeddings.
Both arm blocks and the body block share temporal rotary offsets $0$--$29$, with identity embeddings distinguishing the blocks. Trained by flow matching, the expert predicts a velocity field that covers both arm and body action chunks.}
  \label{fig:overview}
\end{figure}

\subsection{Structurally native visual interface}
\label{sec:principle}

Following NEO~\citep{neo}, \sysname{} adopts a structurally native visual interface that projects raw images directly into the backbone token space through a lightweight two-stage convolutional embedder with an overall stride of $32$, producing one visual token for each $32\times32$ image region. Visual and language tokens are then jointly processed throughout the decoder backbone, while continuous actions are generated by a flow-matching expert, avoiding explicit action tokenization.

Besides, we employ Native Rotary Position Embedding~\citep{neo} to factorize positions into temporal, height, and width coordinates with decoupled rotary frequencies. Note that image tokens retain explicit two-dimensional spatial coordinates, while videos additionally advance along the temporal axis, enabling a unified treatment of images, videos, and interleaved multimodal sequences.

The backbone further adopts modality-aware multimodal attention: text tokens follow causal attention, while visual tokens interact bidirectionally within each visual block. This enables dense spatial modeling while visual and linguistic representations co-evolve across layers. Consequently, the action expert can read from different stages of the visual hierarchy from fine-grained early features to more semantic later representations, rather than being restricted to a single, fixed encoder output.

Notably, we train the 40-layer encoder-free vision-language backbone with 2B parameters through three-stage training. We first establish visual perception through large-scale image-text pretraining while preserving the linguistic capabilities of the pretrained language model. Mid-training then deepens vision-language alignment using diverse captioning, conversation, OCR, and detection data, followed by supervised instruction tuning on high-quality multimodal interactions. This gradually advances the model from multimodal perception to reasoning, yielding a strong foundation for downstream action learning.

\subsection{Multimodal prefix}
\label{sec:visual}

In \paperfigref{fig:overview}, \sysname{} converts robot observations into a unified multimodal prefix composed of previous and current multi-view images, structured language, and robot states. Each temporal block contains up to three camera views, with every view represented by a camera-identity token, projected camera geometry, and patch-level visual tokens. Camera intrinsics and rotations are encoded by lightweight geometry MLPs, while a single patch embedder is shared across all cameras, views, and time steps. Images preserve their native aspect ratios and are dynamically patchified under a fixed budget of up to 88 visual tokens per view, with padded positions masked from attention. This design keeps the visual pathway fully native: pixels enter the language backbone directly, without an external vision transformer, autoencoder, or discrete visual tokenizer.

The visual blocks are followed by a structured prompt specifying the control frequency, robot embodiment, gripper configuration, action-frame reference, and task instruction. Robot proprioception is represented by six additional tokens corresponding to the previous and current states of Arm~0, Arm~1, and the shared body. State MLPs jointly encode the measured values and their validity masks, while additive embeddings identify temporal role and arm/body identity. Components absent from a particular embodiment are simply masked, allowing the same token interface to accommodate different robot configurations.

All prefix tokens are processed jointly by the native backbone using three-dimensional rotary positions over time, height, and width. Visual tokens from the same image share a temporal coordinate, while their spatial coordinates follow the corresponding patch grid; text and state tokens advance along the temporal axis with zero spatial coordinates. Rather than collapsing this computation into a single final visual representation, \sysname{} exposes the layer-wise keys and values of the backbone directly to a lightweight action expert. The expert then combines this multimodal prefix with noisy action tokens and flow time $\tau$, augmented by step, frame, and identity embeddings, to model action chunks for both arms and the body. The three action blocks share temporal rotary offsets over the prediction horizon, while identity embeddings distinguish their semantic roles. Note that the expert predicts a unified velocity field over the continuous action trajectory.

\subsection{Action readout across visual depths}
\label{sec:layermaps}

The action expert reads from a configurable subset of backbone layers. Let $\mathcal{M}=(m_1,\ldots,m_d)$ denote the layer map, where expert layer $j$ attends to the cached keys and values from backbone layer $m_j$. This formulation makes the depth of visual readout an explicit architectural choice as follows:

\begin{itemize}
  \item \texttt{first-28}: $\mathcal{M}=(0,\ldots,27)$, reading only early backbone layers.
  \item \texttt{skip-28}: $28$ layers sampled approximately uniformly across all $40$ backbone layers.
  \item \texttt{end-28}: $\mathcal{M}=(12,\ldots,39)$, reading only late backbone layers.
  \item \texttt{full-40}: all $40$ backbone layers, paired with a $40$-layer action expert.
\end{itemize}

The first three settings use the same 28-layer expert and hidden width, closely matching parameters while isolating action readout. \texttt{full-40} uses deeper experts and is treated separately as a higher-capacity setting.

\begin{table}[t]
\caption{\textbf{Pretraining corpus.}
Nine public dataset families contribute 2{,}231{,}583 episodes and 9{,}927 hours.
\emph{As held} is the share of stored rows; \emph{As sampled} is the share of robot-row draws during training.}
  \label{tab:corpus}
  \centering
  \footnotesize
  \setlength{\tabcolsep}{2.0pt}
  \setlength{\sbarw}{1.0cm}
  \resizebox{\linewidth}{!}{%
  \begin{tabular}{lcrrrrrll}
    \toprule
    & & \multicolumn{5}{c}{Corpus as built}
      & \multicolumn{2}{c}{Share of rows (\%)} \\
    \cmidrule(lr){3-7}\cmidrule(lr){8-9}
    Family & Arm slots & Scenes & Episodes & Rows & Views
      & Hours$^{\dagger}$ & As held & As sampled \\
    \midrule
    AgiBot-World
      & 2 & 198 & 158{,}100 & 282.65M & 594 & 2{,}617.1
      & \sharebar{0.3044}{undcolor}~30.44
      & \sharebar{0.2286}{jointcolor}~22.86 \\
    Bimanual YAM
      & 2 & 1 & 32{,}184 & 75.97M & 3 & 703.5
      & \sharebar{0.0818}{undcolor}~8.18
      & \sharebar{0.1299}{jointcolor}~12.99 \\
    RoboCOIN
      & 2 & 417 & 98{,}797 & 74.05M & 1{,}179 & 679.0
      & \sharebar{0.0798}{undcolor}~7.98
      & \sharebar{0.1285}{jointcolor}~12.85 \\
    Open X-Embodiment
      & 1 & 120 & 1{,}210{,}137 & 30.28M & 154 & 1{,}161.3
      & \sharebar{0.0326}{undcolor}~3.26
      & \sharebar{0.0875}{jointcolor}~8.75 \\
    Galaxea Open-World
      & 2 & 226 & 20{,}911 & 26.73M & 678 & 495.0
      & \sharebar{0.0288}{undcolor}~2.88
      & \sharebar{0.0829}{jointcolor}~8.29 \\
    RoboMIND
      & 1--2 & 18 & 53{,}097 & 19.61M & 33 & ---
      & \sharebar{0.0211}{undcolor}~2.11
      & \sharebar{0.0726}{jointcolor}~7.26 \\
    RH20T
      & 1 & 210 & 32{,}323 & 14.56M & 518 & 426.5
      & \sharebar{0.0157}{undcolor}~1.57
      & \sharebar{0.0639}{jointcolor}~6.39 \\
    DROID
      & 1 & 41 & 40{,}081 & 10.75M & 123 & 199.0
      & \sharebar{0.0116}{undcolor}~1.16
      & \sharebar{0.0560}{jointcolor}~5.60 \\
    InternData-A1 (sim.)
      & 1--2 & 1{,}473 & 585{,}953 & 393.80M & 4{,}162 & 3{,}646.3
      & \sharebar{0.4242}{undcolor}~42.42
      & \sharebar{0.1500}{jointcolor}~15.00 \\
    \midrule
    Total
      & & 2{,}704 & 2{,}231{,}583 & 928.40M & 7{,}444
      & 9{,}927.7$^{\ddagger}$
      & \makebox[\sbarw]{}~100.00
      & \makebox[\sbarw]{}~100.00 \\
    \bottomrule
  \end{tabular}%
  }
\end{table}

\subsection{Continuous action policy modeling}
\label{sec:action}

Following \citet{pi0_2024}, \sysname{} uses a separate action expert to predict continuous action chunks. Each expert layer jointly attends to prefix keys and values cached at its mapped backbone layer and to the action tokens' own keys and values. The prefix $\mathcal{C}$ contains the camera observations and geometry, language instruction, and robot state. During full fine-tuning, the action loss updates both the backbone and the expert.

Each target $\va$ contains an arm chunk in $\mathbb{R}^{2\times30\times26}$ and a shared body chunk in $\mathbb{R}^{30\times8}$. Each arm has 7 joint targets, 3 translation increments, 3 rotation-vector increments, 1 gripper value, and 12 hand-joint targets. The body has 3 base-motion, 3 torso, and 2 head channels. End-effector increments are expressed in centimetres and radians per step, in a camera or robot-base frame. Joint and hand targets are absolute, and body actions use the robot-base frame.
Targets are normalized per embodiment.
Given $\va$, we sample Gaussian noise $\rvepsilon\sim\mathcal{N}(\mathbf{0},\mI)$ and a time $\tau=0.999u$, where $u\sim\mathrm{Beta}(1,1.5)$. On valid dimensions, we construct
\begin{equation}
  \va_\tau=(1-\tau)\rvepsilon+\tau\va.
\end{equation}
The expert predicts the velocity $v_\theta(\va_\tau,\tau,\mathcal{C})$ with target
$\va-\rvepsilon$.
We mask missing steps and channels, and exclude target groups that fall outside the normalization range. The training objective balances four action groups: end-effector motion and gripper, arm joints, hand joints, and body actions, with weights $w_g=(2,1,1,1)$. Let $\ell_g$ denote the mean squared velocity error over valid entries in group $g$, and let $\mathcal{G}$ be the set of groups with valid supervision. We minimize the training objective as follows: 
\begin{equation}
  \mathcal{L}_{\mathrm{action}}
  = \mathbb{E}\left[
      \frac{\sum_{g\in\mathcal{G}} w_g\ell_g}
           {\sum_{g\in\mathcal{G}} w_g}
    \right].
  \label{eq:flow}
\end{equation}
Samples without valid targets contribute zero loss. During inference, we solve the learned flow from $\tau=0$ to $1$ using 10 Euler steps while keeping the prefix KV cache fixed. The resulting action chunk is then de-normalized back to the original action space for execution. We do not apply classifier-free guidance.
\section{Training}
\label{sec:training}

\begin{figure}[t]
  \centering
  \includegraphics[width=\linewidth]{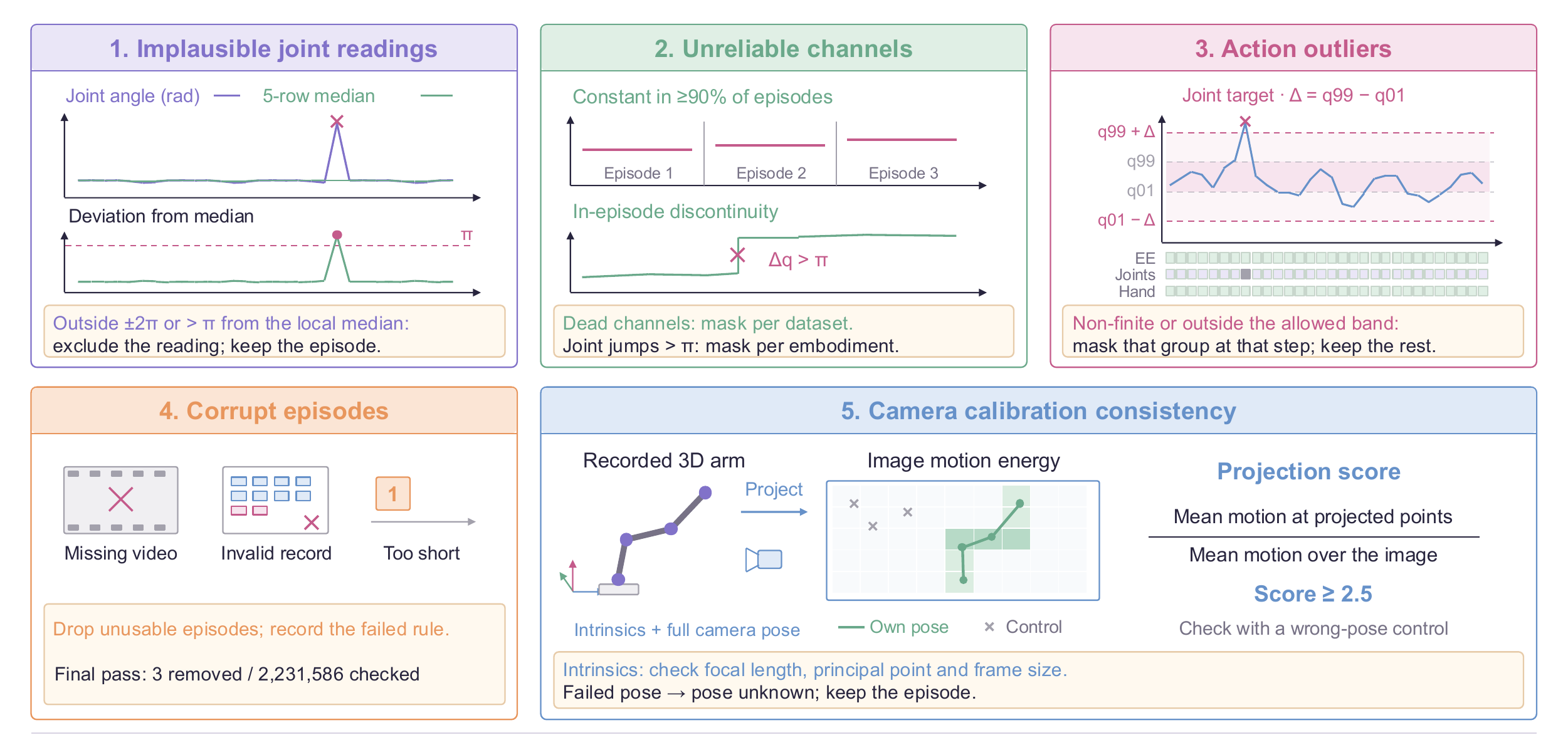}
  \caption{\textbf{Data quality filtering.}
  Illustrative checks cover implausible joint readings, unreliable channels, action outliers, corrupt episodes, and camera calibration, removing only affected data. Joint and projection checks are source-specific; episode retention is determined in the final corpus pass.}
  \label{fig:data_pipeline}
\end{figure}

\subsection{Pretrain Data Overview}
In \papertabref{tab:corpus}, we build the pretrain data from AgiBot-World~\citep{agibotworld2026}, Bimanual YAM~\citep{fang2026molmoact2actionreasoningmodels}, RoboCOIN~\citep{robocoin}, Open X-Embodiment~\citep{open_x_embodiment_rt_x_2023}, Galaxea Open-World~\citep{galaxea2025}, RH20T~\citep{rh20t2023}, RoboMIND~\citep{wu2025robomind}, InternData-A1~\citep{contributors2025internroboticsrepo}, and DROID~\citep{droid2024}, covering diverse embodiments, tasks, and visual conditions.

\subsection{Data Pipeline}
\label{sec:data_pipeline}

We first apply source checks to exclude unreliable measurements and corrupt episodes in \paperfigref{fig:data_pipeline}. Joint
readings outside $\pm 2\pi$, or more than $\pi$ from a five-row local median, are marked invalid. Joint and hand channels that remain constant throughout at least 90\% of a dataset's recorded episodes are masked. Unresolved joint discontinuities larger than $\pi$ are masked for the corresponding embodiment. Action outliers are checked against expanded per-embodiment bounds; an out-of-band or non-finite value masks only its field group at that step, leaving other targets available. Episodes with missing or unreadable data, inconsistent record shapes, invalid recorded values, or too few steps are dropped. 
We also check camera calibration where the required signals are available. Recorded 3D arm positions are projected using the intrinsics and full camera pose. We score image motion at the projected points relative to motion over the image, and compare scores with wrong-pose controls. Untrusted poses are marked unknown while their episodes remain. Intrinsics undergo separate checks for focal length, principal point, field of view and consistency with the stored image size.

\subsection{Training Details}
\label{sec:training_details}

Our training recipe has three progressive stages. In \textit{stage~1}, we freeze the pretrained vision-language backbone and train the action expert and conditioning modules. This stage uses 113{,}400 optimizer steps, a global batch size of 8192, and a peak learning rate of $2\times10^{-4}$. In \textit{stage~2}, we unfreeze the full model and continue training on the same robot pretrain corpus for 75{,}000 steps with the same batch size. The action expert and conditioning modules use a peak learning rate of $2\times10^{-4}$, while the vision-language parameters use $5\times10^{-5}$. We co-train with multi-modal understanding data to retain vision-language perception and reasoning capabilities. Packed understanding examples are added on every other optimizer step, with an auxiliary next-token prediction loss weighted by 0.1 on those steps. In \textit{stage 3}, we continue the full-parameter finetune initialized from stage-2 with our in-house action data for another 75{,}000 steps with the same batch size, all the other settings are consistent with stage-2. Both stages use AdamW with $\beta_1=0.9$, $\beta_2=0.95$, $\epsilon=10^{-8}$ and weight decay $10^{-10}$. Learning rates increase linearly over 2{,}000 warmup steps, then follow cosine decay to 10\% of their respective peaks. We use bfloat16 mixed precision and clip the gradient norm at 1.0.
\section{Experiments}
\label{sec:experiments}

We organize the experiments around four questions. First, what does a native-input backbone bring over encoder-based ones under matched settings (\ref{sec:exp-backbone})? Second, which depths does the action expert read, and what does it extract from them (\ref{sec:exp-depth})? Third, how does retaining multimodal understanding affect action learning (\ref{sec:exp-und})? Finally, how does NEO-VLA perform on in- and out-of-distribution benchmarks (\ref{sec:exp-main})?

\subsection{What the Native-Input Backbone Brings to NEO-VLA?}
\label{sec:exp-backbone}

We compare the native-input backbone of \sysname{} with
Qwen3-VL-2B~\citep{Qwen3-VL} and Gemma-3-4B~\citep{gemma_2025}
to examine how effectively their representations support policy learning. For fair comparisons, all the models are pretrained with Open X-Embodiment for 75k iters under the same setting.
Table~\ref{tab:backbone} reports frozen-backbone training, which updates only the action expert, and full fine-tuning, which updates both the backbone and expert. Results use the final 30{,}000-step checkpoints, with 500 episodes per LIBERO suite. Pretraining configurations and action-expert construction are detailed in the appendix.

\begin{table}[t]
  \caption{\textbf{Backbone comparison on LIBERO.}
  Success rates (\%) at the final 30{,}000-step checkpoint,
  averaged over 2{,}000 episodes.
  Size refers to the backbone, excluding the action expert.
  Frozen updates only the expert; full FT updates the whole stack.
  Native-input rows use the linear-depth layer map.
  Encoder unfrozen updates the vision encoder and action expert with other layers frozen.}
  \label{tab:backbone}
  \centering
  \footnotesize
  \setlength{\tabcolsep}{5pt}
  \resizebox{0.8\linewidth}{!}{%
  \begin{tabular}{@{}l@{\hskip 4pt}cl@{\hskip 5pt}
    S[table-format=2.2]
    S[table-format=2.2]
    S[table-format=2.2]
    S[table-format=2.2]
    S[table-format=2.2]@{}}
    \toprule
    Backbone & Size & Adaptation
      & {Spatial} & {Object} & {Goal} & {Long}
      & {\textbf{Average}} \\
    \midrule

    Qwen3-VL
      & 2B & Frozen
      & 71.4 & 89.8 & 92.8 & 64.0 & 79.5 \\
      & & Full FT
      & 92.2 & 96.6 & 95.8 & 78.6 & 90.8 \\
      & & Encoder unfrozen
      & 91.8 & 92.0 & 94.0 & 79.0 & 89.2 \\
    \midrule

    Gemma-3
      & 4B & Frozen
      & 80.2 & 93.6 & 92.6 & 71.6 & 84.5 \\
      & & Full FT
      & 89.0 & 97.6 & 94.8 & 82.8 & 91.0 \\

    \midrule

    \ours \sysname{}
      & 2B & Frozen
      & 89.2 & 95.8 & 93.2 & 85.4 & 90.9 \\
      \ours & & Full FT
      & \bfseries 95.6 & \bfseries 97.8 & \bfseries 96.0 & \bfseries 92.0 & \bfseries 95.3 \\

    \bottomrule
  \end{tabular}%
  }
\end{table}

The clearest benefit of the native backbone appears before backbone
adaptation. With only its action expert trained, \sysname{} reaches
90.9\% average success, compared with 79.5\% for Qwen3-VL and
84.5\% for Gemma-3. Its frozen-backbone score is also comparable to
both fully fine-tuned encoder-based models. On LIBERO-Long, it reaches
85.4\%, exceeding their full-FT results.
This suggests that the native backbone already provides representations
the expert can use effectively for control, reducing the need to reshape
the backbone during policy learning. Full fine-tuning provides improvements, bringing the average to 95.3\%.

\begin{wraptable}{r}{0.48\textwidth}
  \vspace{-\intextsep}
  \caption{\textbf{Direct encoder access in Qwen3-VL.}
  Average LIBERO success (\%) with the same setting.
  Encoder access adds no parameters.}
  \label{tab:encoder-access}
  \centering
  \footnotesize
  \setlength{\tabcolsep}{2.5pt}
  \resizebox{\linewidth}{!}{%
  \begin{tabular}{@{}l@{\hskip 5pt}
    S[table-format=2.1]
    S[table-format=2.1]@{}}
    \toprule
    Adaptation & {Original interface} & {Encoder access} \\
    \midrule
    Frozen backbone & \bfseries 79.5 & 64.9 \\
    Full FT         & 90.8 & \bfseries 90.9 \\
    \bottomrule
  \end{tabular}%
  }
  \vspace{-\intextsep}
\end{wraptable}

Direct access to visual features alone does not reproduce this advantage.
In a separate Qwen3-VL experiment, we route 13 expert layers to
intermediate vision-tower features. Table~\ref{tab:encoder-access} compares this interface under the same setting. Encoder access lowers frozen-backbone performance from 79.5\% to 64.9\%, while full fine-tuning brings the two interfaces to similar scores, 90.9\% and 90.8\%. These results highlight the importance of how representations are presented to, and aligned with, demonstrating that the action expert only access to the intermediate features is not sufficient. Taken together, the results support a practical contribution of NEO's native backbone: its low-level representations that transfer well to action related tasks naturally, while retaining room for improvement through full fine-tuning.

\shows{NEO's advantage lies not simply in accessing earlier visual features, but in how encoder-free pretraining shapes them into control-ready representations. Its frozen backbone already transfers strongly to policy learning, whereas direct access to intermediate encoder features fails to reproduce this benefit.}

\subsection{Which Backbone Layers Does the Action Expert Use?}
\label{sec:exp-depth}
For a fair comparison, all NEO variants evaluated in this subsection are pretrained on Open X-Embodiment for 75k iterations under identical training settings.

\noindent\textbf{Access to early layers.}
Each expert layer reads cached keys and values from a selected backbone layer. Table~\ref{tab:layermap} compares different choices of these layers. \texttt{skip-28} reads 28 layers spread across the backbone, whereas \texttt{end-28} reads layers 12--39. Both retain all 40 backbone layers and use 28 expert layers. With a frozen backbone, \texttt{skip-28} scores 90.9\% compared with 86.7\% for \texttt{end-28}. It also scores higher with full fine-tuning (95.3\% versus 92.5\%). These comparisons favor keeping early backbone layers accessible. The disjoint \texttt{first-14} and \texttt{last-14} maps show the same trend, but \texttt{first-14} also shortens the backbone. Like \texttt{first-28}, it changes depth and compute as well as readout position.

\begin{table}[t]
\caption{\textbf{Layer-map ablation on LIBERO.}
Success rate (\%) over 2{,}000 rollouts per configuration at training step 30k. \emph{Depth} is the number of backbone layers retained. \texttt{skip-28} and \texttt{end-28} both use a 40-layer backbone and a 28-layer expert. \texttt{first-28} and \texttt{first-14} also shorten the backbone, so the disjoint \texttt{first-14}/\texttt{last-14} comparison changes backbone depth as well as readout position. Higher is better.}
  \label{tab:layermap}
  \centering
  \small
  \setlength{\tabcolsep}{5pt}
  \begin{tabular}{@{}llcccccc@{}}
    \toprule
    \multirow{2}{*}{Adaptation} & \multirow{2}{*}{Layer map}
      & \multirow{2}{*}{Depth}
      & \multicolumn{4}{c}{LIBERO suites $\uparrow$}
      & \multirow{2}{*}{\textbf{Average}} \\
    \cmidrule(lr){4-7}
      & & & Spatial & Object & Goal & Long & \\
    \midrule
    \multirow{3}{*}{\textsc{Frozen}}
      & \texttt{first-28} $[0..27]$    & 28 & 94.8 & 98.4 & 97.6 & 88.8 & \textbf{94.9} \\
      & \texttt{skip-28} $[\mathrm{uniform}]$ & 40 & 89.2 & 95.8 & 93.2 & 85.4 & 90.9 \\
      & \texttt{end-28} $[12..39]$     & 40 & 78.4 & 93.6 & 94.2 & 80.4 & 86.7
 \\
    \midrule
    \multirow{2}{*}{\textsc{Frozen}, disjoint}
      & \texttt{first-14} $[0..13]$    & 14 & 95.6 & 97.2 & 96.2 & 85.4 & \textbf{93.6} \\
      & \texttt{last-14} $[26..39]$    & 40 & 92.2 & 95.2 & 89.4 & 81.8 & 89.6 \\
    \midrule
    \multirow{3}{*}{\textsc{Finetuned}}
      & \texttt{first-28} & 28 & 95.0 & 96.8 & 96.6 & 90.0 & 94.6 \\
      & \texttt{skip-28}  & 40 & 95.6 & 97.8 & 96.0 & 92.0 & \textbf{95.3} \\
      & \texttt{end-28}   & 40 & 94.6 & 95.2 & 95.8 & 84.2 & 92.5\\
    \bottomrule
  \end{tabular}
\end{table}

\begin{figure}[t]
  \centering
  \includegraphics[width=\linewidth]{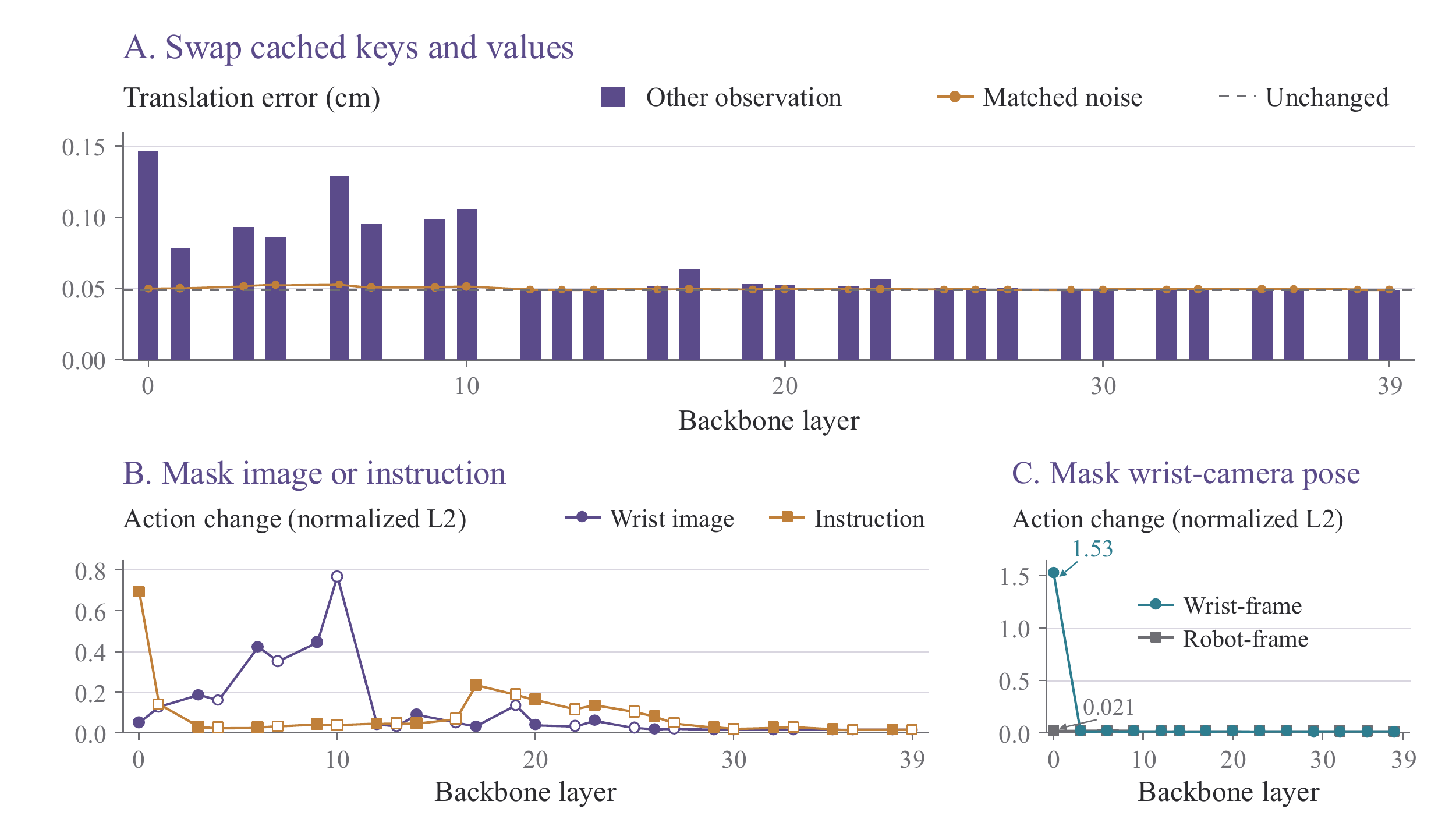}
  \caption{\textbf{Interventions at the expert's attachment points.}
  (A) Translation error after swapping one attachment's prefix keys and
  values with another observation's (purple), adding matched-magnitude
  residual noise (orange), or making no change (gray).
  (B) Action change when masking the wrist image or instruction; filled
  and open markers denote separate even- and odd-attachment runs.
  (C) Wrist-pose masking in the robot-frame model from B and a
  wrist-frame model.}
  \label{fig:layerroles}
\end{figure}

\noindent\textbf{Sensitivity at each attachment point.}
We then intervene at one attachment point at a time on diagnostic \texttt{skip-28} models in \paperfigref{fig:layerroles}. In panel~A, we replace the cached keys and values read by one expert layer with those from another observation and task. Swaps at layers 0--10 increase translation error most. Noise matched to the resulting change in the expert's residual has little effect, staying within 0.004\,cm of the unchanged error. The swap effect therefore depends on the content being replaced, rather than just the size of the disturbance.

Panel~B probes input sensitivity by masking one input span at a single attachment point. The wrist-image response is strongest around layers 6--10, whereas instruction sensitivity peaks at layer~0, with a smaller rise near layer~17. Masking the wrist-camera pose has little effect in the robot-frame model. To confirm that this reflects task dependence rather than probe insensitivity, panel~C evaluates a model predicting actions in the wrist-camera frame, where pose masking produces a strong response at layer~0. Together, these results reveal distinct depth profiles for visual, linguistic, and geometric information.

\shows{Action-relevant information is organized across depth rather than only in the final representation. Early layers dominate frozen transfer, while fine-tuned policies benefit from distributed readouts, with language, appearance, and geometry exerting their strongest influence at different backbone stages.}

\begin{table}[t]
  \caption{\textbf{LIBERO-plus by perturbation axis.}
  \emph{Open} indicates whether both code and weights are publicly available, other results follow \citet{qwenrobotmanip2026}.
  The total is task-weighted rather than the mean of the seven axes;
  \textbf{bold} marks the best value in each column.}
  \label{tab:liberoplus}
  \centering
  \footnotesize
  \setlength{\tabcolsep}{2.5pt}
  \resizebox{\linewidth}{!}{%
  \begin{tabular}{@{}l@{\hskip 4pt}cc@{\hskip 5pt}
    S[table-format=2.1]
    S[table-format=2.1]
    S[table-format=2.1]
    S[table-format=2.1]
    S[table-format=2.1]
    S[table-format=2.1]
    S[table-format=2.1]
    S[table-format=2.1]@{}}
    \toprule
    Model & Size & Open
      & {Camera} & {Robot} & {Language} & {Light}
      & {Background} & {Noise} & {Layout}
      & {\textbf{Total}} \\
    \midrule

    $\pi_0$
      & 3B & \cmark
      & 13.8 & 6.0 & 58.8 & 85.0
      & 81.4 & 79.0 & 68.9 & 53.6 \\

    StarVLA
      & 4B & \cmark
      & 52.5 & 49.8 & 88.5 & 95.7
      & 95.7 & 73.0 & 76.9 & 74.1 \\

    ABot-M0
      & 4B & \cmark
      & 60.4 & 67.9 & 86.4 & 96.2
      & 91.6 & 86.4 & 82.6 & 80.5 \\

    Cosmos-Policy
      & 2B & \cmark
      & 75.8 & 63.3 & 81.7 & 96.5
      & 88.9 & 92.7 & 82.2 & 82.2 \\

    $\pi_{0.5}$
      & 3B & \cmark
      & 78.4 & 73.6 & 80.8 & 96.2
      & 94.1 & 89.0 & 84.5 & 84.4 \\

    \midrule

    Qwen-RobotManip-scratch
      & 4B & \pmark
      & 70.4 & 44.9 & 88.1 & 95.8
      & 95.5 & 84.4 & 79.1 & 78.3 \\

    Being-H0.7
      & 3B & \pmark
      & 82.0 & 59.0 & 82.8 & \bfseries 97.8
      & 90.0 & 93.5 & \bfseries 88.5 & 84.8 \\

    Qwen-RobotManip
      & 4B & \pmark
      & 87.2 & \bfseries 75.5 & 85.6 & 96.6
      & \bfseries 97.7 & \bfseries 97.7
      & 87.3 & 89.0 \\

    \midrule

    \ours \sysname{}
      & 2B & \cmark
      & \bfseries 94.5 & 72.6 & \bfseries 91.7 & 96.2
      & 96.6 & 96.9 & 82.4 & \bfseries 89.7 \\

    \bottomrule
  \end{tabular}%
  }
\end{table}

\subsection{How Multi-modal understanding affect action policy training?}
\label{sec:exp-und}

\begin{wraptable}{r}{0.4\textwidth}
  \vspace{-14pt}
\caption{\textbf{Preserving multimodal understanding enables robust VLA post-training.} 
Understanding scores (\%) and RoboTwin 2.0 clean-to-randomized (C2R) success (\%) before post-training (Baseline), after co-training with diverse understanding corpus (+ Co-training), and after action-only fine-tuning (+ Action-only).
The action-only model cannot produce valid text, while co-training keeps understanding capability and gives the best manipulation policy.}
  \label{tab:recipe}
  \centering
  \footnotesize
  \setlength{\tabcolsep}{1.8pt}
  \resizebox{\linewidth}{!}{%
  \begin{tabular}{@{}lccc@{}}
    \toprule
    & Baseline & + Co-training & + Action-only \\
    \midrule
    MMB  & \bfseries 82.7 & 81.2 & -- \\
    RWQA & 66.8 & \bfseries 68.4 & -- \\
    AI2D        & \bfseries 82.9 & 82.3 & -- \\
    TextVQA     & \bfseries 76.6 & 76.4 & -- \\
    ChartQA    & \bfseries 80.8 & 80.4 & -- \\
    \midrule
    RoboTwin C2R & -- & \textbf{69.4} & 10.7 \\
    \bottomrule
  \end{tabular}
  }
  \vspace{-60pt}
\end{wraptable}

VLA post-training typically either fine-tunes on action data alone or co-trains with general understanding data.
Table~\ref{tab:recipe} compares the two against the checkpoint before post-training. Co-training with common understanding corpora keeps
understanding within 1.5 points of the baseline on MMBench-EN~\citep{MMBench}, RealWorldQA~\citep{xai_realworldqa_2024},
AI2D~\citep{kembhavi2016diagram}, TextVQA~\citep{singh2019towards} and ChartQA~\citep{masry2022chartqa}, and reaches 69.4\% on RoboTwin~2.0
clean-to-randomized (C2R). Action-only fine-tuning loses the ability to
produce valid text, and its policy falls
to 10.7\% on C2R. 
Co-training therefore preserves the backbone's
understanding and gives the strongest policy under scene randomization.

\shows{Multimodal understanding and robust action are not independent capabilities. Preserving the backbone's pretrained understanding during post-training leads to substantially stronger policy generalization, while eroding it causes action robustness to collapse.}

\subsection{Standard Benchmark Results}
\label{sec:exp-main}
\begin{wraptable}{r}{0.48\textwidth}
  \vspace{-14pt}
  \caption{\textbf{LIBERO}.
  NEO-VLA is evaluated over 2{,}000 episodes per suite
  at step 27{,}500.
  \emph{Size} denotes parameter scale.
  \emph{Open} indicates public weights.}
  \label{tab:libero}
  \centering
  \scriptsize
  \setlength{\tabcolsep}{1.8pt}
  \resizebox{\linewidth}{!}{%
  \begin{tabular}{@{}l@{\hskip 3pt}cc@{\hskip 4pt}
    S[table-format=3.1]
    S[table-format=3.1]
    S[table-format=3.1]
    S[table-format=2.1]
    S[table-format=2.2]@{}}
    \toprule
    Model & Size & Open
      & {Spatial} & {Object} & {Goal} & {Long}
      & {\textbf{Average}} \\
    \midrule

    OpenVLA
      & 7B & \cmark
      & 84.7 & 88.4 & 79.2 & 53.7 & 76.5 \\

    $\pi_{0.5}$
      & 3B & \cmark
      & 98.8 & 98.2 & 98.0 & 92.4 & 96.8 \\

    MMaDA-VLA
      & 8B & \cmark
      & 98.8 & 99.8 & 98.0 & 95.2 & 98.0 \\

    EO-1
      & 3B & \cmark
      & \bfseries 99.7 & \bfseries 99.8 & \bfseries 99.2 & 94.8 & 98.4 \\

    Qwen-RobotManip
      & 4B & \pmark
      & {---} & {---} & {---} & {---}
      &\textbf{99.1} \\

    \midrule

    \ours \sysname{}
      & 2B & \cmark
      & 98.4	& 99.4	& 95.6 & \bfseries 96.2 & 97.4 \\

    \bottomrule
  \end{tabular}%
  }
\end{wraptable}
Simulator versions, camera setups, training data, and evaluation protocols differ across prior work, so these results should be viewed as benchmark context rather than controlled comparisons.

\textbf{LIBERO}~\citep{libero}. In \papertabref{tab:libero}, \sysname{} reaches an average success rate of 97.4.
Here EO-1 and MMaDA-VLA obtain similarly high scores, at 98.4 and 98.0,
respectively, while Qwen-RobotManip reports a single aggregate score of 99.1
without a per-suite breakdown.
The small differences among the strongest methods make LIBERO increasingly
difficult to use for distinguishing current policies.
We therefore report it mainly to show that \sysname{} retains strong
in-distribution performance rather than to draw conclusions from small
cross-paper differences.

\textbf{LIBERO-Plus}~\citep{liberoplus} provides a more informative view by evaluating seven types of distribution shift separately. As shown in \papertabref{tab:liberoplus}, \sysname{} achieves an overall score of 89.7, the highest among models in the table with both code and weights publicly available. Its robustness is particularly strong under camera and language perturbations, reaching 94.5 and 91.7, respectively. The gains are less pronounced under robot and layout changes, which alter the embodiment or scene configuration rather than the visual observation itself. This suggests that \sysname{} is especially robust to changes in how the scene is perceived, rather than to all forms of distribution shift. This pattern is consistent with the layer analysis in \papersecref{sec:exp-depth}, where the action expert relies heavily on early visual representations.

\begin{wraptable}{r}{0.48\textwidth}
  \caption{\textbf{LIBERO-Pro.}
  Evaluation of VLA models under Object, Position,
  Semantic, and Task perturbations.
  \emph{Size} denotes parameter scale.
  }
  \label{tab:liberopro}
  \centering
  \scriptsize
  \setlength{\tabcolsep}{2pt}
  \resizebox{\linewidth}{!}{%
  \begin{tabular}{@{}l@{\hskip 4pt}c@{\hskip 5pt}
    S[table-format=2.1]
    S[table-format=2.1]
    S[table-format=2.1]
    S[table-format=2.1]
    S[table-format=2.1]@{}}
    \toprule
    Model & Size & {Object} & {Position}
      & {Semantic} & {Task} & {\textbf{Avg.}} \\
    \midrule

    OpenVLA
      & 7B & 93.0 & 0.0 & 97.3 & 0.0 & 47.6 \\

    $\pi_0$
      & 3B & 90.5 & 0.0 & 90.5 & 0.0 & 45.3 \\

    $\pi_{0.5}$
      & 3B & \bfseries 96.0 & \bfseries 20.8
      & 95.8 & 0.8 & \bfseries 53.3 \\

    X-VLA
      & 0.9B & 79.0 & 0.8 & 96.8 & \bfseries 6.8 & 45.8 \\

    \midrule

    \ours \sysname{}
      & 2B & 84.3 & 3.0 & \bfseries 98.0 & 6.3 & 47.9 \\

    \bottomrule
  \end{tabular}%
  }
\end{wraptable}

\paragraph{LIBERO-Pro}~\citep{zhou2025liberopro} probes generalization under four distinct perturbations: object appearance, object position, instruction semantics, and task goals. As shown in Table~\ref{tab:liberopro}, NEO-VLA performs particularly well under semantic variation, achieving 98.0, the highest score in the table. It also remains competitive under object changes, but degrades substantially when object positions or task goals are altered. This contrast is informative: NEO-VLA is robust to changes in visual appearance and instruction wording, yet still struggles when successful execution requires adapting to unfamiliar spatial configurations or action goals. Together with the LIBERO-Plus results, this suggests that the native backbone provides strong perceptual and semantic robustness, while flexible spatial and task-level adaptation remains a key challenge.

\begin{wraptable}{r}{0.47\textwidth}
  \vspace{-\intextsep}
  \caption{\textbf{RoboTwin 2.0.}
  We report the \emph{official setting}
  with 2.5K episodes.
  \emph{Size} denotes parameter scale.
  \emph{Open} indicates public weights.}
  \label{tab:robotwin}
  \centering
  \scriptsize
  \setlength{\tabcolsep}{2.5pt}
  \resizebox{\linewidth}{!}{%
  \begin{tabular}{@{}l@{\hskip 4pt}cc@{\hskip 5pt}
    S[table-format=2.2]
    S[table-format=2.2]
    S[table-format=2.2]@{}}
    \toprule
    Model & Size & Open & {Clean} & {Clean2Rand} \\
    \midrule

    FastWAM
      & 6B & \cmark & 77.80  & 1.90 \\

    StarVLA-OFT
      & 3B & \cmark & 46.52  & 3.16 \\

    $\pi_{0.5}$
      & 3B & \cmark & 70.70  & 46.00 \\

    OLA-Sem
      & 8B & \pmark & 75.12  & 67.56 \\

    Qwen-RobotManip
      & 4B & \pmark & \bfseries 89.58  & 68.12 \\

    \ours \sysname{}
      & 2B & \cmark & 75.78  & \bfseries 69.40 \\
    \bottomrule
  \end{tabular}%
  }
\end{wraptable}

\paragraph{RoboTwin 2.0}~\citep{robotwin2} further evaluates robustness to distribution shift through randomized environments in \papertabref{tab:robotwin}.
We report results under the official 2.5K-episode setting. NEO-VLA achieves 75.78 on the clean split and retains 69.40 under Clean2Rand evaluation. Although its clean performance trails the strongest closed models, it exhibits only a modest degradation under randomization, indicating that the learned policy transfers reliably beyond the training distribution. Among methods with both training code and model weights publicly available, we can discover that NEO-VLA achieves the strongest Clean2Rand result. This result suggests that the native backbone does more than support strong in-distribution manipulation control: it also preserves stable visual representations and action predictions as the environment changes, leading to stronger robustness under distribution shift.

\subsection{Real-World Evaluation}

\subsubsection{Real-World Setup}

\begin{figure}[t]
  \centering
  \includegraphics[width=\textwidth]{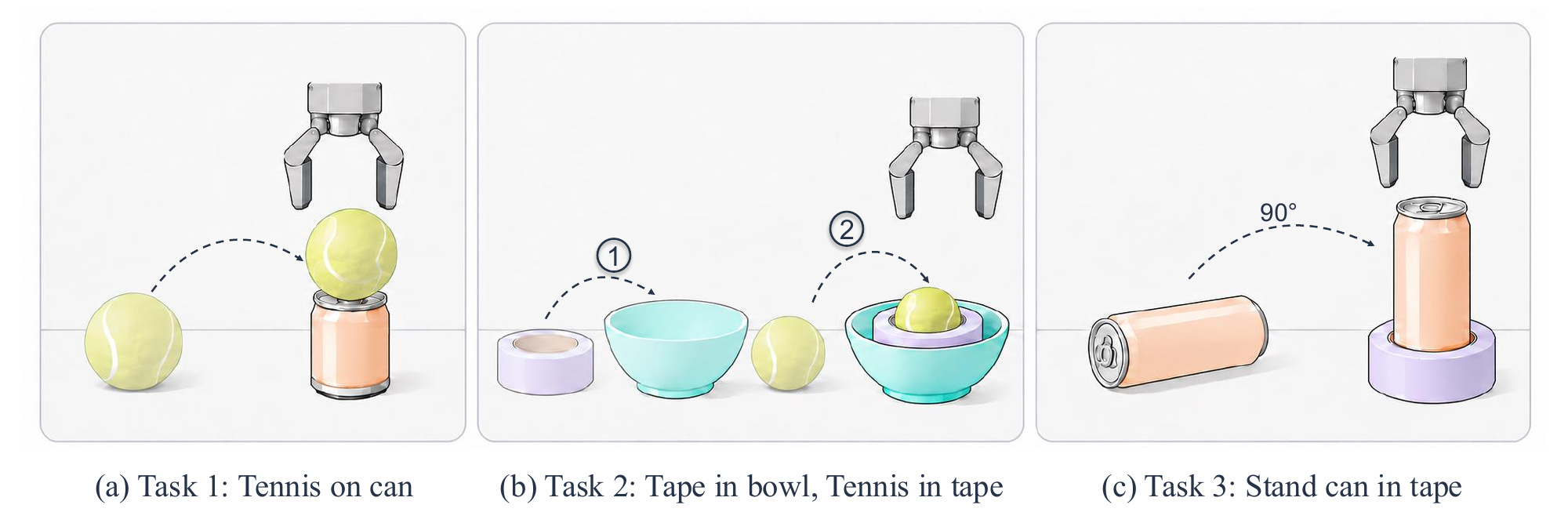}
\caption{\textbf{Real-world manipulation benchmark.}
(a) Place a tennis ball on a can.
(b) Place a tape roll in a bowl, then place a tennis ball inside the roll.
(c) Reorient a horizontal can and insert it upright into a tape roll.
The tasks test placement precision, multi-step execution, and object reorientation. Each model is fine-tuned on 50 episodes per scene and evaluated over 20 trials.}
\label{fig:realworld_tasks}
  \label{fig:real_data}
\end{figure}

To evaluate real-world manipulation comprehensively, we construct a small benchmark with three tasks, as illustrated in Figure~\ref{fig:real_data}: (a) pick up a tennis ball and place it on top of a can; (b) pick up a roll of sellotape and place it into a bowl, then pick up the tennis ball and place it inside the sellotape; and (c) pick up a horizontally placed can and insert it into the sellotape. These tasks are designed to test both fine-grained manipulation precision and long-horizon execution. For each task, models are fine-tuned on only 50 demonstration episodes and evaluated over 20 trials, including 10 in-domain trials and 10 out-of-domain trials.

\begin{wraptable}{r}{0.40\textwidth}
  \vspace{-\intextsep}
  \caption{\textbf{Real-World evluation.}
  We report task success rate over 20 episodes per task, with 10 trials in domain and the other 10 trials out-of-domain.}
  \label{tab:real_data}
  \centering
  \scriptsize
  \setlength{\tabcolsep}{2.5pt}
  \resizebox{0.9\linewidth}{!}{%
  \begin{tabular}{@{}l@{\hskip 4pt}cccc@{\hskip 5pt}
    S[table-format=2.2]
    S[table-format=2.2]
    S[table-format=2.2]@{}}
    \toprule
    Model & Size & Task1 & Task2 & Task3 \\
    \midrule
    StarVLA
      & 3B & 1/20 & 2/20  & 0/20 \\

    $\pi_{0.5}$
      & 3B & 10/20 & 14/20  & 13/20 \\

    \ours \sysname{}
      & 2B & 12/20 & 17/20 & 13/20 \\
    \bottomrule
  \end{tabular}%
  }
\end{wraptable}

\begin{figure}[t]
  \centering
  \includegraphics[width=\textwidth]{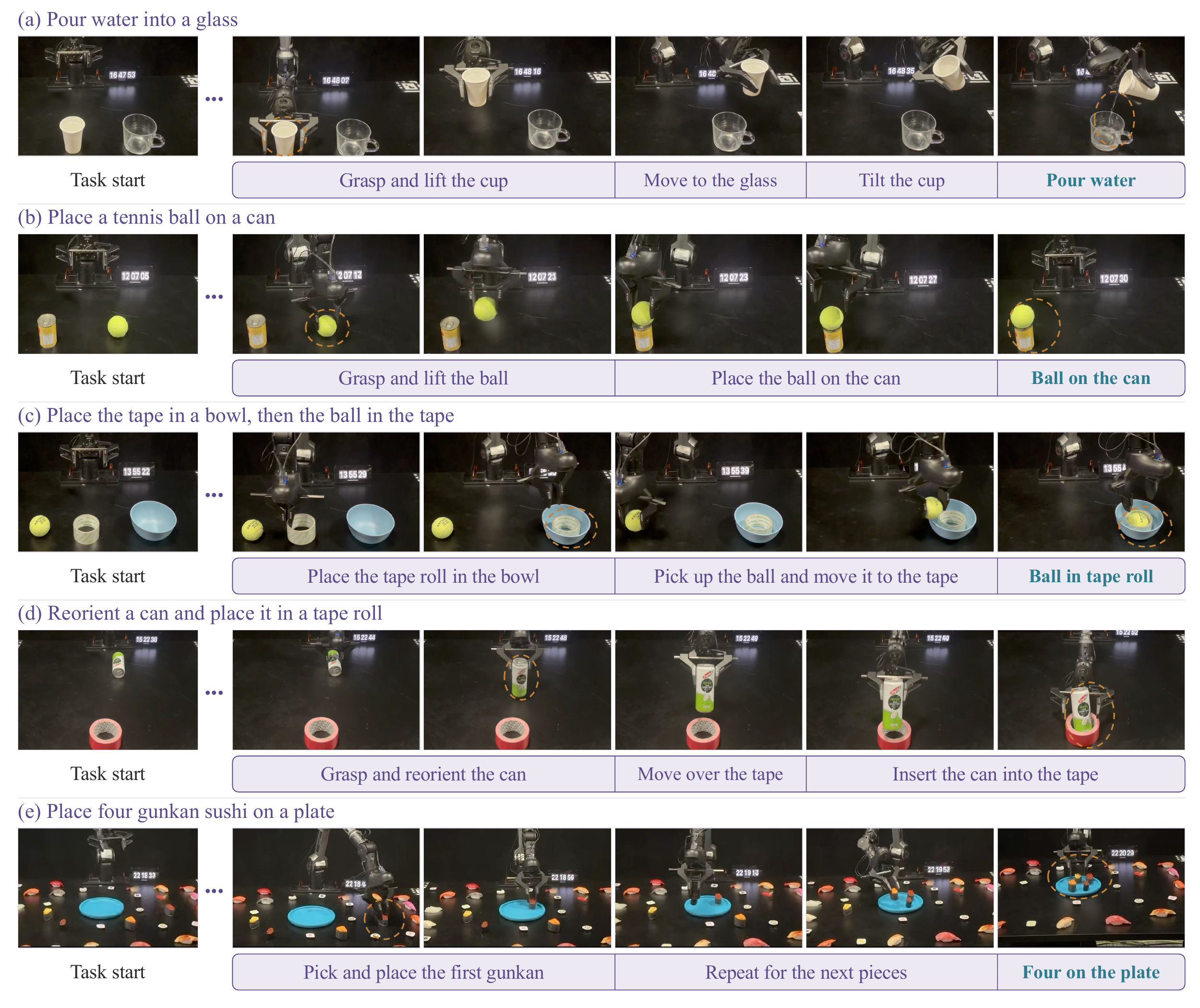}
\caption{\textbf{Real-world NEO-VLA demonstrations.}
(a) Pick up a paper cup and pour water into a glass.
(b) Place a tennis ball on a can.
(c) Place a tape roll in a bowl, then place the ball inside the roll.
(d) Reorient a horizontal can and insert it upright into a tape roll.
(e) Pick four gunkan sushi pieces, one at a time, from a table of mixed sushi and place them on a plate.
Frames are arranged chronologically from left to right, with dashed circles highlighting key interaction moments and critical manipulation steps.}
\label{fig:real_policy}
\end{figure}

\subsubsection{Evaluation Results}
As shown in Table~\ref{tab:real_data}, we compare NEO-VLA with $\pi_{0.5}$~\citep{pi05_2025} and StarVLA~\citep{community2026starvla} using the same real-world training data and each method's recommended post-training setup. Task 1 and Task 3 emphasize precise object interaction and placement, while Task 2 requires sustained long-horizon execution across multiple sequential manipulation steps. NEO-VLA achieves the strongest overall performance after the same fine-tuning process, with success rates of 12/20, 17/20, and 13/20 across three real-world tasks. It substantially outperforms StarVLA and matches or exceeds $\pi_{0.5}$ in every setting, despite its smaller model scale.
More importantly, NEO-VLA maintains strong performance when object configurations deviate from those seen during training, indicating that the learned policy captures transferable visual-action structure rather than simply memorizing demonstrated layouts. The qualitative examples in Figure~\ref{fig:real_policy} further illustrate this capability across diverse interaction patterns, including precise placement, object reorientation, sequential manipulation, and multi-step execution. Taken together, these results show that the native VLA design transfers effectively to real-world control, supporting both fine-grained manipulation and long-horizon behavior under limited real-world supervision, while retaining robustness to changes in the observed scene.

\subsubsection{Inference Latency}
\begin{wraptable}{r}{0.40\textwidth}
  \vspace{-\intextsep}
  \caption{\textbf{Inference latency summary.}
  We report milliseconds per query on one RTX 3090 in BF16, with raw 30-step chunks.}
  \label{tab:latency}
  \centering
  \scriptsize
  \setlength{\tabcolsep}{2.5pt}
  \resizebox{0.9\linewidth}{!}{%
  \begin{tabular}{@{}l@{\hskip 8pt}
    S[table-format=3.1]
    S[table-format=3.1]@{}}
    \toprule
    Stage & {10 steps} & {2 steps} \\
    \midrule
    \multicolumn{3}{@{}l}{\textit{Policy}} \\
    Patch embedder              & 3.1   & 2.7   \\
    Backbone prefix (40 layers) & 84.8  & 84.7  \\
    Action expert sampling      & 163.4 & 32.4  \\
    Forward pass, eager         & 515.4 & 210.7 \\
    Forward pass, compiled      & 282.5 & 147.8 \\
    \midrule
    \multicolumn{3}{@{}l}{\textit{Robot (compiled)}} \\
    Request $\rightarrow$ action chunk    & 261.9 & 153.5 \\
    Request $\rightarrow$ executable plan & 286.2 & 178.5 \\
    \bottomrule
  \end{tabular}%
  }
\end{wraptable}

We also measure latency on the real robot (pika), with the policy served on a single NVIDIA RTX~3090 (24\,GB) in bf16. Each query returns one raw 30-step action chunk, which we execute as predicted without real-time chunking~\citep{black2025real}. Table~\ref{tab:latency} reports our default of 10 Euler steps and, for reference, a 2-step setting. 
With 10 steps, a compiled forward pass takes 282.5\,ms. Most of it goes to the action expert, whose 10 Euler steps take 163.4\,ms. The backbone prefix pass takes 84.8\,ms and runs only once per query, since every step reuses its key--value cache. The convolutional patch embedder costs 3.1ms. Compiling the backbone and expert layers with \texttt{torch.compile}, together with caching the conditioning inputs within each request, cuts the forward pass from 515.4 to 282.5ms and changes predicted positions by at most 0.002cm. On the robot, a query takes 261.9\,ms from request to the returned chunk, and 286.2ms until an executable plan is ready after inverse kinematics. Cutting sampling to 2 steps shortens the expert to 32.4ms and the full query to 178.5ms, so the prefix pass becomes the main cost. Fewer steps also change the sampled actions; note that all success rates in Table~\ref{tab:real_data} use 10 steps.

\section{Conclusion}
\label{sec:conclusion}

NEO-VLA challenges a common assumption in modern VLA design: that action should operate on visual representations already distilled for semantic understanding. Our results instead suggest that effective control benefits from retaining access to the visual computation before it collapses into a final semantic representation. A native backbone is already highly transferable to action when frozen, early-layer access is consistently valuable, and simply exposing an encoder-based VLM to intermediate visual features does not recover the same benefit. This indicates that what matters is not encoder removal alone, but how visual representations are formed and exposed to the policy throughout the backbone.
The broader picture is equally revealing. NEO-VLA is highly robust when the appearance or observation of a scene changes, yet remains challenged when the spatial configuration or task goal itself changes. This asymmetry suggests that the next bottleneck in embodied intelligence may no longer be perception alone, but converting perceptual understanding into flexible action under changing geometry and intent. We therefore view native VLA modeling as a step toward representations that are not only semantically meaningful, but intrinsically useful for control.

\bibliography{iclr2027_conference}
\bibliographystyle{iclr2027_conference}


\end{document}